\documentclass{article}

\PassOptionsToPackage{numbers, sort&compress}{natbib}
\usepackage{wrapfig}

\usepackage[preprint]{neurips_2026}

\usepackage[utf8]{inputenc}
\usepackage[T1]{fontenc}
\usepackage{hyperref}
\usepackage{url}
\usepackage{booktabs}
\usepackage{amsfonts}
\usepackage{nicefrac}
\usepackage{microtype}
\usepackage{xcolor}
\usepackage{graphicx}
\usepackage{amsmath}
\usepackage{array}
\usepackage{algorithm}
\usepackage{algorithmic}
\usepackage{amsthm}
  \newtheorem{remark}{Remark}
\title{Boosting RL-Based Visual Reasoning with Selective Adversarial Entropy Intervention}

\author{
  Yang Yu\textsuperscript{1} \quad
  Zhuangzhuang Chen\textsuperscript{1} \quad
  Lanqing Li\textsuperscript{2} \quad
  Xiaomeng Li\textsuperscript{1*}\\
  \textsuperscript{1} The Hong Kong University of Science and Technology\\
  \textsuperscript{2} The Chinese University of Hong Kong\\
  \texttt{eeyangyu@ust.hk, eexmli@ust.hk}
}

\begin{document}

\maketitle

\begin{abstract}

  Recently, reinforcement learning (RL) has become a common approach for enhancing the reasoning capabilities of vision-language models (VLMs). Considering existing RL-based finetuning methods, entropy intervention has emerged as an effective way to improve exploration, thereby improving policy performance. Notably, most existing studies intervene in entropy by simply controlling the update of specific tokens during policy optimization of RL. However, they largely ignore entropy intervention during rollout sampling, which can improve GRPO by increasing
  response diversity. In this paper, from a maximum-entropy RL perspective, we propose Selective-adversarial Entropy Intervention (SaEI), a sampling-stage approximation to entropy-regularized exploration. It locally increases selective token entropy and expands the support of rollout trajectories without changing the GRPO policy-update objective. Specifically, we first enhance policy entropy by adversarially perturbing the visual input along the gradient of a trajectory-entropy objective. Then, we analyze how adversarial perturbations interact with token advantages, and propose Token-selective Entropy Computation (TsEC) to improve the effectiveness of adversarial sampling via a bilateral entropy bound. Extensive experiments on both in-domain and out-of-domain datasets show that our method improves policy exploration through entropy intervention and enhances visual reasoning performance.
  
\end{abstract}

\section{Introduction}
\label{sec:intro}

Reinforcement learning (RL) advances large language models (LLMs) by enhancing their reasoning capabilities, enabling planning, reflection, and generalization beyond mere memorization \cite{guo2025deepseek, jaech2024openai, seed2025seed1}.
Motivated by this fact, RL-based post-training has been explored to boost vision language models (VLMs), which equip LLMs with vision encoders, to solve complex tasks such as mathematical reasoning and code generation \cite{fan2025posterior, ekbote2025murphy}.

\begin{figure}[t]
    \centering
    \begin{minipage}[t]{0.48\textwidth}
        \centering
        \includegraphics[width=\linewidth]{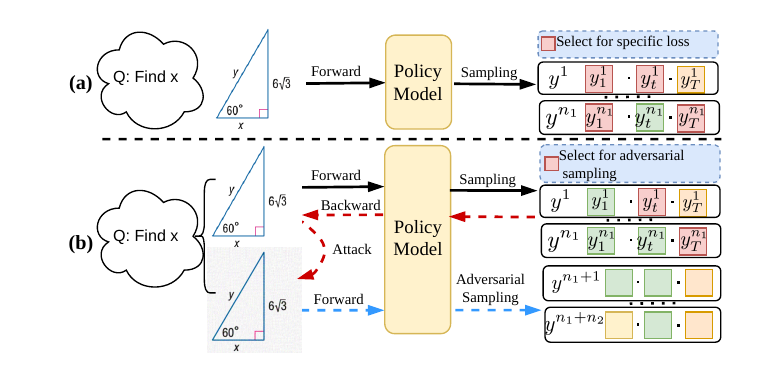}
        \vspace{-4mm}
     	\caption{(a) Existing methods intervene in policy entropy by controlling the update of specific tokens. (b) Our method utilizes entropy-guided adversarial samples to intervene from the perspective of RL sampling. Here, $y^i$ represents a response, while $y^i_t$ stands for a token in $y^i$.}
	\label{intro}
    \end{minipage}
    \hfill
    \begin{minipage}[t]{0.48\textwidth}
        \centering
        \includegraphics[width=\linewidth]{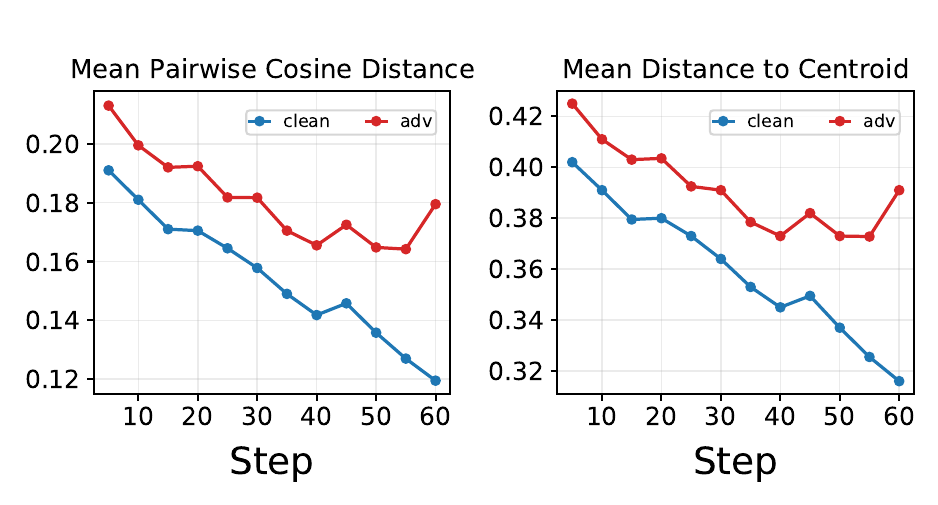}
        \vspace{-4mm}
	\caption{Comparison of response diversity between clean and adversarial images across sampling steps. At each
  step, we sample 512 prompts and generate 10 responses per prompt, measuring diversity by the mean pairwise
  cosine distance and the mean distance to the centroid. }
	\label{diversity}
    \end{minipage}
\end{figure}

Group relative policy optimization (GRPO) \cite{guo2025deepseek} has attracted extensive attention, owing to its simplicity and scalability with competitive results. It allows LLMs to learn from a group of their own outputs by estimating advantages in a group-based manner. Notably, the diversity of their outputs is determined by policy exploration capacity, which can be quantified by policy entropy during rollout sampling. 
Meanwhile, recent studies \cite{yu2025dapo, cui2025entropy} argue that GRPO cannot maintain a sufficient level of policy entropy during LLM fine-tuning, resulting in a performance plateau. To address this problem, entropy intervention has been proposed to improve exploratory ability. As shown in Fig.\ref{intro} (a), most existing works \cite{cui2025entropy,chen2026flexible, cheng2026reasoning} achieve entropy intervention by controlling the update of specific tokens during the policy optimization of RL.
Although entropy intervention during policy optimization has been well explored, existing works overlook entropy intervention during RL sampling that can boost the performance of GRPO by improving the diversity of responses.

Intuitively, entropy intervention during RL sampling can be achieved by randomly perturbing the image-question pair. However, random perturbation lacks an explicit optimization target and therefore does not provide a controllable way to increase rollout entropy. Moreover, we empirically find that the performance of random perturbation is sensitive to the perturbation strength, as shown in Fig. 4. This observation raises a natural question: \emph{\textbf{how can we perform entropy intervention during rollout sampling in a principled and controllable manner?}} We answer this question from a maximum-entropy reinforcement learning \cite{haarnoja2018soft} perspective. In token-level visual reasoning, a decoding trajectory can be viewed as a sequence of states $h_t=(I,Q,y_{<t})$ and actions $y_t$. Maximum-entropy RL augments reward maximization of an action with policy entropy, encouraging the rollout policy to maintain broader sampling support instead of prematurely concentrating on a narrow response distribution. In GRPO, this is useful because group-relative learning relies on responses sampled from the rollout policy, and the diversity of these samples affects the information available for policy optimization. Rather than modifying the GRPO policy-update objective, we aim to approximate maximum-entropy exploration at the sampling stage.

Based on this view, we propose Selective-adversarial Entropy Intervention (SaEI), which consists of entropy-guided adversarial sampling (EgAS) and token-selective entropy computation (TsEC). The key idea of EgAS is to treat token entropy as an adversarial objective and perturb the visual input along the entropy-ascent direction. Since the adversarial gradient is computed from policy entropy, the resulting adversarial visual context explicitly encourages the old rollout policy to sample from a broader response distribution. As shown in Fig.~\ref{diversity}, adversarial images produce more diverse responses than clean images. This distinguishes SaEI from random visual perturbation: random noise may increase or decrease entropy unpredictably, whereas EgAS directly targets selective token entropy. Then, we analyze the relationship between adversarial perturbations and token advantages, and propose \textbf{TsEC} to maximize the effectiveness of EgAS via the bilateral entropy bound.

To the best of our knowledge, we are the first to use adversarial visual contexts to intervene in policy entropy during RL sampling for VLMs.
Our main contributions can be summarized as follows:
\begin{itemize}
\item We formulate entropy-guided adversarial sampling as a sampling-stage approximation to
      maximum-entropy exploration in token-level visual reasoning. This view provides a principled
      explanation for increasing rollout diversity during GRPO sampling.
\item We propose entropy-guided adversarial sampling (EgAS), which treats token entropy
      as an adversarial objective and perturbs the visual input to generate higher-diversity rollouts for
      policy optimization.

\item We propose token-selective entropy computation (TsEC), which selects informative tokens according to the interaction between adversarial perturbations and token advantages.

\item Extensive experiments on in-domain and
      out-of-domain datasets demonstrate that SaEI improves
      policy exploration and consistently enhances visual reasoning performance over strong
      GRPO-based baselines.
  \end{itemize}

\section{Related Works}

Large Language Models (LLMs) \cite{team2024qwen2, dubey2024llama, achiam2023gpt} have achieved great success in processing language information. Furthermore, VLMs \cite{bai2025qwen2, wang2024qwen2, chen2024expanding, liu2024improved} integrate the vision encoder into LLMs so that they can handle vision modalities such as images and videos. 
Some studies \cite{besta2024graph, xu2025llava} suggest that reasoning techniques can greatly improve the performance of both LLMs and VLMs in solving complex problems. Therefore, reasoning tasks have attracted intensive research in this area.
Researchers have explored several approaches to stimulate the reasoning capacity of LLMs and VLMs. A line of research \cite{wu2025grounded, muennighoff2025s1} performs SFT on pre-collected long CoT data which follows specific steps, leading to improvement of reasoning capacity. The CoT data can be collected by guiding LLMs to generate structured reasoning templates—such as first descriptions and then conclusions \cite{xu2025llava, thawakar2025llamav}. Other works generate the CoT data using advanced search methods such as Beam Search \cite{wu2024v, yao2023tree}, and Monte Carlo Tree Search (MCTS) \cite{xie2024monte, yao2024mulberry}. 
However, SFT-based reasoning approaches suffer from two main challenges. First, it is costly to collect massive high-quality CoT data. Second, SFT-based reasoning models merely memorize the structured reasoning format and knowledge within CoT, struggling to adapt to unseen tasks \cite{chu2025sft}. Recently, reinforcement learning (RL) has emerged as another effective approach to activate reasoning capacity \cite{jaech2024openai, seed2025seed1, guo2025deepseek}.

DeepSeek-R1 \cite{guo2025deepseek} proposes a new value-model-free RL algorithm, Group Relative Reinforcement Learning (GRPO), which achieves great success in improving the reasoning capacity with massive verifiable questions. 
Inspired by the success of Deepseek-R1, a lot of works \cite{meng2025mm, huang2025vision, feng2025video} utilize GRPO to enhance the general reasoning capacity of VLMs. These works first build a large-scale multimodal dataset consisting of verifiable visual questions. They then deploy LLMs like GPT-4o to generate high-quality CoTs for a part of these questions. The questions with CoTs can be used for SFT, enabling MLLMs to imitate reasoning behavior in CoTs. Finally, they study how to combine SFT and RL to improve the reasoning capacity of MLLMs. Besides visual conversation tasks, some other works demonstrate that RL can enhance reasoning in visual perception tasks, like referring object segmentation \cite{liu2025seg}, object counting \cite{liu2025visionreasoner}, and open vocabulary detection \cite{liu2025visual}.

Some studies on RL mechanisms have focused on several limitations of GRPO like entropy collapse \cite{cui2025entropy}, training instability \cite{yu2025dapo}, reward bias \cite{liu2025understanding}, insufficient policy exploration \cite{liu2025noisyrollout}, etc. Entropy Mechanism \cite{cui2025entropy} notices the entropy collapse of  reinforcement learning for Reasoning Language Models. It further proposes to mitigate entropy collapse by restricting the update of tokens with high covariances between advantage and action probability. \cite{cheng2026reasoning} modifies the GRPO policy-update objective by directly adding entropy bonus to advantage to encourage trajectories with higher entropy.
In our work, we propose a sampling-stage approximation to entropy-regularized exploration. Rather than modifying the GRPO policy-update objective, we approximate maximum-entropy exploration during rollout generation. This design decouples exploration from optimization: we increase the entropy of the behavior policy at the sampling stage, thereby producing more diverse candidate reasoning paths, while the subsequent policy update remains governed by the original task reward. As a result, we improve exploration and advantage estimation without introducing an explicit entropy bonus that may bias or destabilize optimization.

\begin{figure*}[t]
	\centering
	\includegraphics[width=0.97\textwidth]{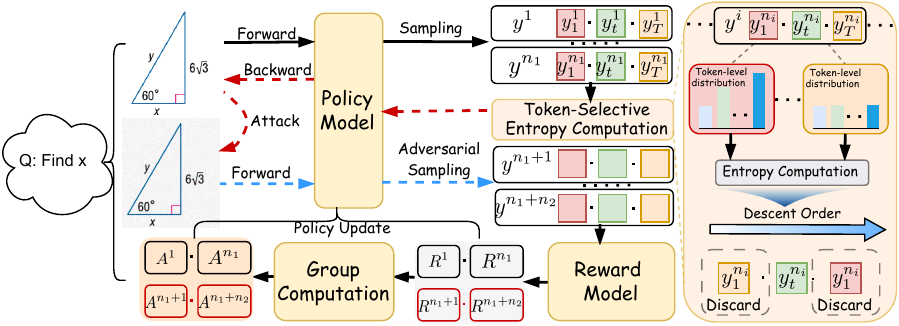}
	\caption{Overview of our proposed Selective adversarial Entropy Intervention (SaEI). It consists of entropy-guided adversarial sampling (EgAS) and token-selective entropy computation (TsEC). EgAS intervenes in policy entropy from the perspective of RL sampling, by taking entropy as an adversarial objective to disrupt visual input with the corresponding gradient. TsEC discards tokens with the highest and lowest entropy, while keeping tokens with moderate entropy for the entropy computation of the adversarial objective.}
	\label{overview}
\end{figure*}

\section{Methods}

\subsection{Preliminaries}
\paragraph{Token-level RL formulation.} 
Given an image-question pair $q = (I, Q)$, a response $y=(y_1,\ldots,y_T)$ is generated autoregressively. At decoding step \(t\), we regard the reasoning
history $h_t=(I,Q,y_{<t})$ as the state and the next token $a_t=y_t$ as the action. A vision-language model induces a stochastic policy
\[
  \pi_\theta(y_t \mid h_t).
\]
GRPO utilizes the old model $\pi_{\theta_{old}}$ to generate a group of $n$ responses $\{y^i\}_{i=1}^n$. The rule-based reward function then gives group rewards $\{R^i\}_{i=1}^n$ by evaluating these responses. Next, all tokens or actions in a response share the same advantage $\tilde{A}^i_t$ which is estimated by normalizing the group rewards:
\begin{equation}
\label{grpo_adv}
\tilde{A}^i_t
 = \frac{R^i-mean(\{R^i\}_{i=1}^n)}{std(\{R^i\}_{i=1}^n)}.
\end{equation}
The key role of rollout sampling is therefore to produce a diverse group of candidate trajectories
so that more information can be available for policy optimization.

\subsection{Entropy-guided Adversarial Sampling (EgAS)}
  \label{sec:egas}

  \paragraph{Maximum-entropy objective.}
  Maximum-entropy reinforcement learning, as used in soft actor-critic
  \citep{haarnoja2018soft}, augments reward maximization with an entropy term $H\!\left(\pi(\cdot \mid h_t)\right)$.
  In a token-level reasoning problem, the analogous objective can be written as
  \begin{equation}
  \label{eq:maxent_objective}
  J_{\rm ME}(\pi)
  =
  \mathbb{E}_{q,y\sim\pi}
  \left[
  R(q,y)
  -
  \beta {\rm D_{\mathrm{KL}}}\!\left(\pi(\cdot \mid q)\,\|\,\pi_{\rm ref}(\cdot \mid q)\right)
  +
  \alpha \sum_{t=1}^{T} H\!\left(\pi(\cdot \mid h_t)\right)
  \right],
  \end{equation}
  where $\alpha>0$ controls the strength of entropy maximization and $\beta>0$
  controls the reference-policy constraint. The entropy term prevents premature
  policy collapse and encourages the policy to preserve multiple near-optimal
  reasoning modes. In this sense, entropy is not merely a numerical regularizer: it directly controls the exploration capacity of the rollout policy. 
  However, explicitly adding the entropy term with a fixed $\alpha$ would bias or destabilize policy optimization. Instead, we propose an entropy-guided adversarial sampling that achieves maximum-entropy exploration in an implicit manner. In return, it allows to produce more diverse candidate reasoning paths while keeping the policy update governed by the original task reward.

  \paragraph{GRPO with EgAS.} As shown in Fig.\ref{overview}, SaEI introduces EgAS to encourage policy exploration in RL sampling. The old policy model
  $\pi_{\theta_{\rm old}}$ first generates a group of $n$ responses
  $\{y^i\}_{i=1}^{n}$ for the clean image-question pair
  $q_{\rm cle}=(I,Q)$. We calculate the average token-wise entropy to quantify the rollout policy entropy:
  \begin{equation}
  \label{eq:rollout_entropy}
  \begin{aligned}
  H\!\left(\pi_{\theta_{\rm old}},\{y^i\}_{i=1}^{n}\right)
  &=
  -\mathbb{E}_{\{y^i\}_{i=1}^{n},\pi_{\theta_{\rm old}}}
  \left[
  \log \pi_{\theta_{\rm old}}\!\left(y_t^i \mid q_{\rm cle},y_{<t}^i\right)
  \right] \\
  &=
  -\frac{1}{n}\sum_{i=1}^{n}\frac{1}{|y^i|}
  \sum_{t=1}^{|y^i|}
  \mathbb{E}_{y_t^i\sim\pi_{\theta_{\rm old}}}
  \left[
  \log \pi_{\theta_{\rm old}}\!\left(y_t^i \mid q_{\rm cle},y_{<t}^i\right)
  \right].
  \end{aligned}
  \end{equation}

  Intuitively, we can intervene in policy entropy by randomly perturbing the
  image-question pair $q_{\rm cle}$. However, random perturbation lacks an
  explicit relation to entropy intervention and therefore cannot precisely
  control the direction and degree of exploration. Adversarial learning provides
  a way to connect an adversarial objective with an input perturbation through
  back-propagation. We therefore use the policy entropy in
  Eq.~\eqref{eq:rollout_entropy} as the adversarial objective and back-propagate
  its gradient to the input. Since $q_{\rm cle}=(I,Q)$ consists of an image $I$ and a
  textual question $Q$, and visual inputs are continuous and high-dimensional, we
  apply the adversarial perturbation to the visual input while keeping the textual
  question unchanged.

  In adversarial RL sampling, we first generate $n_1$ clean responses
  $\{y^i\}_{i=1}^{n_1}$ based on the clean image $I$. We then compute
  $H(\pi_{\theta_{\rm old}},\{y^i\}_{i=1}^{n_1})$ with
  Eq.~\eqref{eq:rollout_entropy}. This entropy is used as the adversarial
  objective to be enlarged. Specifically, projected gradient ascent is applied to
  obtain the adversarial image:
  \begin{equation}
  \label{eq:egas_pgd}
  I_{\rm adv}^{k+1}
  =
  \Pi_{\mathcal{B}_{\epsilon_{\rm adv}}(I)}
  \left(
  I_{\rm adv}^{k}
  +
  \eta\,
  {\rm sign}
  \left(
  \nabla_{I_{\rm adv}^{k}}
  H\!\left(\pi_{\theta_{\rm old}},\{y^i\}_{i=1}^{n_1}\right)
  \right)
  \right),
  \end{equation}
  where $k$ indexes the PGD iteration, $\eta>0$ is the adversarial step size,
  $\Pi_{\mathcal{B}_{\epsilon_{\rm adv}}(I)}$ projects the image into the
  $\epsilon_{\rm adv}$-bounded neighborhood of the clean image, and
  ${\rm sign}(\cdot)$ is the element-wise sign function. By ascending the entropy
  gradient, the obtained $I_{\rm adv}$ encourages policy exploration with
  increased token-level entropy.

  With the adversarial image $I_{\rm adv}$, we generate another group of $n_2$
  responses $\{y^i\}_{i=1}^{n_2}$ using the adversarial image-question pair
  $q_{\rm adv}=(I_{\rm adv},Q)$. We then mix the clean and adversarial responses
  to form
  $\{y^i\}_{i=1}^{n_1+n_2}
  =
  \{\{y^i\}_{i=1}^{n_1}\cup\{y^i\}_{i=1}^{n_2}\}$.
  The corresponding rewards are
  $\{R^i\}_{i=1}^{n_1+n_2}
  =
  \{\{R^i\}_{i=1}^{n_1}\cup\{R^i\}_{i=1}^{n_2}\}$.
  The token advantages are estimated using Eq. \ref{grpo_adv}. The policy model $\pi_{\theta}$ is optimized by maximizing the PPO-style objective:
  \begin{equation}
  \begin{aligned}
  \label{grpo_our}
  & \mathcal{J}(\theta) = \mathbb{E}_{[q_{cle} \sim P_D, \{ \{\mathbf{y}_i\}_{i=1}^{n_1} \sim
  \pi_{\theta_{\text{old}}}(\cdot|q_{cle}) \cup \{\mathbf{y}_i\}_{i=1}^{n_2} \sim \pi_{\theta_{\text{old}}}
  (\cdot|q_{adv})\}]}
  \\
  &  \frac{1}{{n_1+n_2}} \sum_{i=1}^{n_1+n_2} \frac{1}{|y^i|} \sum_{t=1}^{|y^i|} \Big \{ \min
  \Big[ \frac{\pi_{\theta}(y_t^i|q, y_{<t}^i)}{\pi_{\theta_{\text{old}}}(y_t^i|q , y_{<t}^i)} \tilde{A}^i_t,
  \text{clip}\Big(\frac{\pi_{\theta}(y_t^i|q, y_{<t}^i)}{\pi_{\theta_{\text{old}}}(y_t^i|q, y_{<t}^i)},
  1-\epsilon, 1+\epsilon \Big) \tilde{A}^i_t  \Big] - \beta D_{kl}  \Big \},
  \end{aligned}
  \end{equation}
  We compute the likelihoods under the current and old policies using the same
  image-question pair $q \in \{q_{\rm cle}, q_{\rm adv}\}$.
  Our method therefore modifies
  only the RL sampling stage while keeping the policy-update objective unchanged. 
  
{\paragraph{Theoretical insight.} Herein, we provide a theoretical analysis showing that SaEI approximates the effect of maximum-entropy reinforcement learning through entropy intervention at the sampling stage, without explicitly adding the entropy term in Eq.~\eqref{eq:maxent_objective}.} For brevity, let $F(I)=H\!\left(\pi(\cdot \mid h_t)\right)$ denote the selected token entropy
  used as the adversarial objective. For
  $u=\operatorname{sign}(\nabla_I F(I))$ and a small perturbation
  $I_\epsilon=I+\epsilon u$, EgAS locally increases rollout entropy:
  \begin{equation}
  \label{eq:entropy_ascent_main}
  F(I_\epsilon)-F(I)
  =
  \epsilon\|\nabla_I F(I)\|_1
  +
  O(\epsilon^2).
  \end{equation}
  This local entropy ascent also induces a corresponding change in the rollout
  policy. At a selected decoding state $h_t$, let $z_t(a)$ denote the old-policy
  logit of token $a$ and let
  $p_{0,t}(a)=\operatorname{softmax}(z_t)(a)$ be the corresponding old token
  distribution. Under a first-order logit expansion,
  $z_{t,\epsilon}(a)=z_t(a)+\epsilon b_t(a)+O(\epsilon^2)$, the adversarial
  rollout policy becomes
  \begin{equation}
  \label{eq:egas_tilt_main}
  p_{t,\epsilon}(a)
  =
  \frac{
  p_{0,t}(a)\exp(\epsilon b_t(a))
  }{
  \mathbb{E}_{a'\sim p_{0,t}}[\exp(\epsilon b_t(a'))]
  }
  +
  O(\epsilon^2).
  \end{equation}
 $p_{t,\epsilon}(a)$ can be viewed as a KL-regularized exponentiated policy tilt of the old policy, and connects EgAS to soft policy improvement. We interpret this adversarially tilted rollout policy from the perspective of maximum-entropy exploration.
 In the reward-free case, the closed-form maximum-entropy policy at state $h_t$ is
  $\operatorname{softmax}(\frac{\beta}{\beta+\alpha}z_t)$, which flattens the old
  logits and increases token entropy. For small $\alpha$, its first-order entropy
  gain satisfies
  \begin{equation}
  \label{eq:maxent_entropy_gain_state}
  H\!\left(\operatorname{softmax}\!\left(\frac{\beta}{\beta+\alpha}z_t\right)\right)
  -
  H\!\left(\operatorname{softmax}(z_t)\right)
  =
  \frac{\alpha}{\beta}
  \operatorname{Var}_{a\sim p_{0,t}}[z_t(a)]
  +
  O(\alpha^2).
  \end{equation}
  Let $\mathcal{S}_{\gamma}$ be the selected token set retained by TsEC, and define $V_{\mathcal{S}}
  =
  \frac{1}{|\mathcal{S}_{\gamma}|}
  \sum_{t\in\mathcal{S}_{\gamma}}
  \operatorname{Var}_{a\sim p_{0,t}}[z_t(a)].$
  Averaged over selected tokens, a maximum-entropy coefficient $\alpha$ therefore
  induces a first-order entropy gain of approximately
  $\frac{\alpha}{\beta}V_{\mathcal{S}}$. Matching this gain with the EgAS entropy
  gain in Eq.~\eqref{eq:entropy_ascent_main} gives
  \begin{equation}
  \label{eq:entropy_gain_alpha_main}
  \alpha_{\rm eff}
  \approx
  \beta\epsilon
  \frac{
  \|\nabla_I F(I)\|_1
  }{
  V_{\mathcal{S}}
  },
  \qquad
  V_{\mathcal{S}}>0.
  \end{equation}

  This establishes a quantitative connection between the step size $\epsilon$ of entropy-guided adversarial sampling and effective $\alpha$ of maximum-entropy exploration in Eq.~\eqref{eq:maxent_objective}. Specifically,
  Eq.~\eqref{eq:entropy_gain_alpha_main} suggests that the effective
  maximum-entropy coefficient induced by EgAS is sample-dependent. For a fixed
  adversarial step size $\epsilon$, $\alpha_{\rm eff}$ is larger when the selected
  entropy objective has a larger input-space gradient $\|\nabla_I F(I)\|_1$, and
  smaller when the selected-token logit variance $V_{\mathcal{S}}$ is larger.
  Therefore, EgAS adaptively applies stronger entropy regularization to samples
  whose selected token entropy is more responsive to visual perturbation.
  Meanwhile, the clean-adversarial rollout mixture expands the support of sampled
  trajectories without adding an explicit entropy bonus to the GRPO update. The
  detailed derivation is provided in Appendix~\ref{app:maxent_egas}.

\subsection{Token-Selective Entropy Computation (TsEC)}

Our adversarial sampling is guided by the policy entropy that is averaged over all token positions. As illustrated by \cite{wang2025beyond}, tokens with different entropy levels play different roles in reasoning: the \emph{lowest-entropy} tokens primarily complete linguistic structures with factual knowledge memorized by LLMs, while the \emph{highest-entropy} tokens function as pivotal decision points that determine the direction of reasoning trajectories. This suggests that not every token is equally useful when constructing the adversarial objective.
We start by examining how adversarial perturbation interacts with the advantage sign of the tokens. The analysis is conducted along two complementary directions: (i) tokens with positive advantage, and (ii) tokens with negative advantage.

\paragraph{Positive-advantage tokens: an upper entropy bound.} 
A positive-advantage token belongs to a relatively successful trajectory. Meanwhile, high-entropy positions often indicate pivotal decision points where the model is uncertain among multiple reasoning branches. Considering these factors, our adversarial sampling for those high-entropy positive-advantage tokens offers little additional exploration and may destabilize the correct reasoning path. Thus, we derive an upper entropy bound on positive-advantage tokens to preserve successful decisions while still allowing diversity to be encouraged at less fragile positions.

\paragraph{Negative-advantage tokens: a lower entropy bound.}
A negative-advantage token belongs to a relatively unsuccessful trajectory. Our adversarial sampling aims to produce diverse samples and, more importantly, to potentially redirect incorrect responses toward correct reasoning paths. In this regard,  high-entropy negative tokens mark uncertain decision points where the model may have committed to an incorrect branch and are therefore more likely to be corrected under perturbation. In contrast, those low-entropy negative tokens typically correspond to deterministic lexical or structural components, whose perturbation has limited influence on the reasoning direction. To this end, we derive a lower entropy bound on negative-advantage tokens.

\paragraph{Bilateral entropy bound.} 
To ease the burden of explicitly judging the sign of advantages during the sampling stage, we apply both bounds simultaneously to every trajectory. The upper bound removes extremely high-entropy tokens, protecting successful pivotal decisions, while the
  lower bound removes extremely low-entropy tokens, avoiding ineffective perturbations on deterministic
  scaffolding. As illustrated in Fig.~\ref{overview}, the adversarial objective is computed only over tokens selected by this bilateral bound.

\begin{equation}
\label{grpo_entropy_selective}
\begin{aligned}
&\mathcal{H}(\pi_{\theta_{old}}, \{y^i\}_{i=1}^{n_1} ) = - \frac{1}{n_1} \sum_{i=1}^{n_1} \frac{1}{|\hat{y}^i|} \sum_{t=1}^{|\hat{y}^i|} \mathbb{E}_{\hat{y}_t^i \sim \pi_{\theta_{old}}} [\log \pi_{\theta_{\text{old}}}(\hat{y}_t^i|q, y_{<t}^i) ], \\
 &
 \text{s.t.} \quad  \gamma * |\hat{y}^i|<\textbf{rank}(\hat{y}_t^i) < (1- \gamma) *  |\hat{y}^i|.
 \end{aligned}
\end{equation}

\section{Experiments}
In this section, we first introduce the datasets, benchmarks and baseline methods. Then we present and analyze the comparison with baseline methods. Lastly, we conduct ablation studies to analyze our method in detail.

\subsection{Datasets and Benchmarks}

Our experiments are conducted with two mathematics-related datasets: Geometry3K \cite{lu2021inter} and MM-Eureka \cite{meng2025mm}. Geometry3K is a dataset focusing on geometric problems, consisting of 2.1K training samples and 0.6K test samples. MM-Eureka is a multimodal mathematical K-12 level dataset, composed of 15.6K training samples and 2K test samples. We consider the two test sets from Geometry3K and MM-Eureka as in-domain test sets. Additionally, given the well-known generalization ability of RL, we evaluate our model on the following two out-of-domain  benchmarks: MathVision \cite{wang2024measuring}, and HallusionBench \cite{guan2024hallusionbench}. We also report the average accuracy of these two benchmarks to get an overall evaluation.

\subsection{Baseline Methods}
We compare our method with vanilla GRPO \cite{guo2025deepseek}, KL-Cov \cite{cui2025entropy}, Entropy Adv. \cite{cheng2026reasoning}, Entropy Con. \cite{chen2026flexible}, and NoisyRollout \cite{liu2025noisyrollout}. KL-Cov is designed to intervene in policy entropy by controlling the update of tokens with high covariances between the action probability and the corresponding advantage value during policy optimization of RL. Entropy Adv. adds the entropy value of each token to its advantage value so that the entropy can be enlarged during RL optimization. Entropy Con. proposes a dynamic clipping threshold to precisely manage entropy during RL optimization. NoisyRollout tries to improve GRPO by adding Gaussian noise to visual input and controlling noise strength with an annealing strategy. For both datasets, vanilla GRPO and KL-Cov use the group size of $n=12$, while NoisyRollout and our SaEI use the group size of $n_1=6, n_2=6$ for training. For a fair comparison, we reproduce vanilla GRPO, NoisyRollout, and KL-Cov following the default settings and general hyperparameters of EasyR1, which is the same as our SaEI.

\begin{table*}[ht]
\begin{center}
\caption{Comparison with existing methods on MM-Eureka. We report the mean and standard deviation over three runs. All results are reported in percentages ($\%$). $\dagger$ indicates that the model is not finetuned. Vanilla GRPO and KL-Cov use the group size of $n=12$, while NoisyRollout and our SaEI use the group size of $n_1=n_2=6$, resulting in a total group size of $n=12$ for fair comparison.}
\resizebox{0.99\textwidth}{!}{ 
\begin{tabular}{lcccc}
  \toprule[2pt]
  Method                 & MM-Eureka                 & HallusionBench                 & MathVision   & OOD
  Avg.          \\
  \toprule[1pt]
  Qwen2.5-VL-7B-Instruct $\dagger$
  & 41.70
  & 64.00
  & 25.95
  & 44.98                \\
  \toprule[1pt]
  Vanilla GRPO (Nature'25) \cite{guo2025deepseek}
  & 62.45 $\pm$ 1.57
  & 70.48 $\pm$ 0.43
  & 27.50 $\pm$ 0.74
  & 48.99      \\

  NoisyRollout (NeurIPS'25) \cite{liu2025noisyrollout}
  & 62.93 $\pm$ 0.08
  & \underline{71.33 $\pm$ 0.61}
  & 27.96 $\pm$ 0.12
  & \underline{49.65}    \\

  KL-Cov \cite{cui2025entropy}
  & \underline{63.34 $\pm$ 4.51}
  & 70.67 $\pm$ 1.07
  & \underline{28.60 $\pm$ 0.82}
  & 49.54     \\

  Entropy Adv. (AAAI'26) \cite{cheng2026reasoning}
  & 62.82 $\pm$ 0.76
  & 71.29 $\pm$ 0.21
  & 26.89 $\pm$ 0.43
  & 49.09 \\

  Entropy Con. \cite{chen2026flexible}
  & 62.87 $\pm$ 0.13
  & 70.42 $\pm$ 0.87
  & 28.38 $\pm$ 0.71
  & 49.40 \\

  \toprule[1pt]
  SaEI (Ours)
  & \textbf{64.45 $\pm$ 1.26}
  & \textbf{71.85 $\pm$ 0.46}
  & \textbf{29.21 $\pm$ 0.32}
  & \textbf{50.53} \\
  \toprule[2pt]
  \end{tabular}}
  \vspace{-3mm}
\label{tab:mm-eureka}
\end{center}
\end{table*}

\begin{table*}[ht]
\begin{center}
\caption{Comparison with existing methods on Geometry3K. We report the mean and standard deviation over three runs. The group size is the same as Tab.\ref{tab:mm-eureka}. The best results are marked in bold, and the second-best results are underlined.}
\resizebox{0.99\textwidth}{!}{
\begin{tabular}{lcccc}
  \toprule[2pt]
  Method                 & Geometry3K                & HallusionBench                 & MathVision   & OOD
  Avg.             \\
  \toprule[1pt]
  Qwen2.5-VL-7B-Instruct$\dagger$
  & 39.27
  & 64.00
  & 25.95
  & 44.98
  \\
  \toprule[1pt]
  Vanilla GRPO (Nature'25) \cite{guo2025deepseek}
  & 54.02 $\pm$ 0.08
  & 70.18 $\pm$ 0.46
  & 27.61 $\pm$ 0.23
  & 48.90      \\

  NoisyRollout (NeurIPS'25) \cite{liu2025noisyrollout}
  & 54.74 $\pm$ 1.26
  & 70.11 $\pm$ 0.45
  & 27.70 $\pm$ 0.35
  & \underline{48.91}      \\

  KL-Cov \cite{cui2025entropy}
  & \underline{55.91 $\pm$ 0.17}
  & 68.31 $\pm$ 3.07
  & \underline{27.92 $\pm$ 0.33}
  & 48.12      \\

  Entropy Adv. (AAAI'26) \cite{cheng2026reasoning}
  & 55.02 $\pm$ 1.06
  & 70.03 $\pm$ 1.17
  & 26.19 $\pm$ 0.17
  & 48.11 \\

  Entropy Con. \cite{chen2026flexible}
  & 54.85 $\pm$ 0.51
  & \underline{70.21 $\pm$ 0.95}
  & 27.13 $\pm$ 0.51
  & 48.67 \\

  \toprule[1pt]
  SaEI (Ours)
  & \textbf{56.18 $\pm$ 0.51}
  & \textbf{70.38 $\pm$ 0.36}
  & \textbf{28.17 $\pm$ 0.47}
  & \textbf{49.28} \\
  \toprule[2pt]
  \end{tabular}}
\label{tab:geo3k}
\end{center}
\end{table*}

\subsection{Implementation Details}
We implement our method based on a widely used EasyR1 \cite{zheng2025easyr1} codebase. We choose Qwen2.5-VL-7B-Instruct as our base model, because it demonstrates robust foundational abilities, making it an ideal candidate for RL finetuning. We use the default settings and hyperparameters from EasyR1 to focus on the development of our algorithm. Therefore, we keep vision encoder trainable during training. We use a rollout temperature of 1.0, a rollout batch size of 512, a global batch size of 128, a KL loss weight $\beta$ of 1e-2, and a learning rate of 1e-6. Our policy loss is computed using the $token\_mean$ average mode. For our adversarial learning, the iteration number of the PGD attack $T$ is set to 1. The step size of adversarial attack $\eta_{\mathrm{adv}}$ is set to $\frac{2}{255}$ and $\frac{3}{255}$ for MM-Eureka and Geometry3K, respectively. We run 60 steps for Geometry3K and 90 steps for MM-Eureka. We run each experiment three times and report its average and standard deviation. For evaluation, we again follow the default setting of EasyR1 to set the temperature to 0.6 and report the average pass@1 accuracy. Our experiments are conducted using 8 H100 GPUs with 80G memory. The prompts used are presented in appendix.

\subsection{Comparison with Other Methods}
As shown in Tab.~\ref{tab:mm-eureka}, after training on the MM-Eureka dataset, the model trained using our method attains an accuracy of 64.45$\%$ on the in-domain test set, demonstrating a 2.00 percentage points improvement over the model trained with vanilla GRPO. Our method outperforms NoisyRollout with a margin of 1.52 percentage points, which demonstrates entropy-guided adversarial images are more effective than randomly perturbed images for improving policy exploration. Although intervening in policy entropy from different perspectives, our method also outperforms KL-Cov with a gap of 1.11$\%$. The superiority of our method is further emphasized by its performance on out-of-domain benchmarks. Our method achieves a 1.37$\%$ improvement over the model trained with vanilla GRPO, and outperforms KL-Cov with 1.18 percentage points on HallusionBench. Our method also achieves the best results in terms of ``OOD Avg.".

As shown in Tab.~\ref{tab:geo3k}, our method again improves the in-domain accuracy by 2.16 percentage points and slightly surpasses the second-best
method, KL-Cov, by 0.27 percentage points. While other methods struggle to improve the performance on OOD benchmarks, our method can consistently enhance the performance on OOD benchmarks. 
As shown in Tab.\ref{geosk_n81}, we conduct comparison on Geometry3K with a smaller group size $n=8$. The policy exploration capacity decreases as the group size becomes smaller. Compared with KL-Cov, our method can better improve policy exploration under the smaller group size, with an improvement over vanilla GRPO of 2.55 percentage points vs. KL-Cov's improvement over vanilla GRPO of 1.39 percentage points.

\begin{table}[t]
  \centering

  \begin{minipage}{0.59\linewidth}
  \centering
  \caption{Comparison with KL-Cov on Geometry3K with the group size of 8.}
  \label{geosk_n81}
  \resizebox{\linewidth}{!}{
  \begin{tabular}{ccc}
  \hline
  Method & Group Size & Geometry3K \\ \hline
  Vanilla GRPO & $n=8$ & 52.47$\pm$0.97 \\
  KL-Cov & $n=8$ & 53.86$\pm$0.53 \\
  SaEI (Ours) & $n=8\ (n_1=n_2=4)$ & \textbf{55.02$\pm$0.10} \\ \hline
  \end{tabular}
  }
  \end{minipage}
  \hfill
  \begin{minipage}{0.38\linewidth}
  \centering
  \caption{Component analysis of our SaEI. }
  \label{mainab}
  \resizebox{\linewidth}{!}{
  \begin{tabular}{cc|c}
  \hline
  EgAS & TsEC & Geometry3K \\ \hline
       &      & 52.47$\pm$0.97 \\
  $\checkmark$ &      & 53.86$\pm$0.53 \\
  $\checkmark$ & $\checkmark$ & \textbf{55.02$\pm$0.10} \\ \hline
  \end{tabular}
  }
  \end{minipage}

  \end{table}

\begin{figure}[t]
    \centering
    \begin{minipage}[t]{0.38\textwidth}
        \centering
        \includegraphics[width=\linewidth]{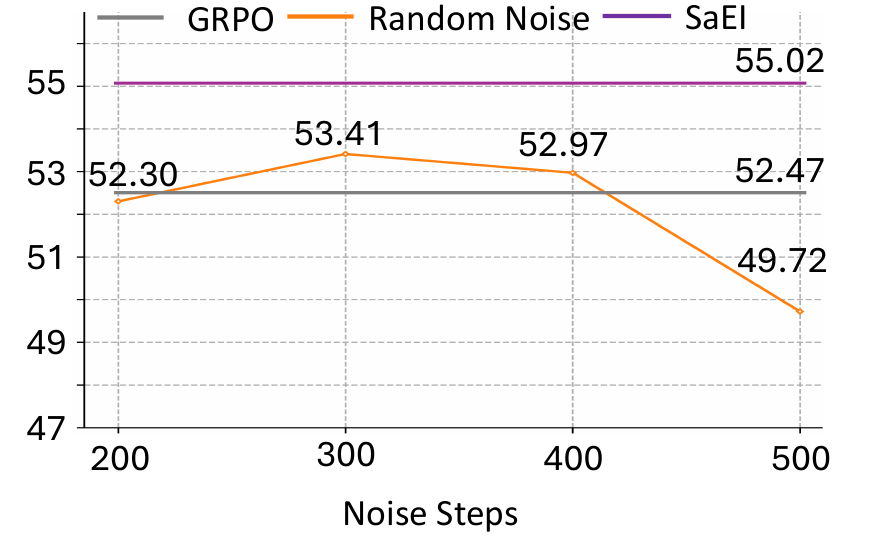}
     	\caption{Comparison with random noise, where multiple-step Gaussian noise is added to the image.}
	\label{random_noise}
    \end{minipage}
    \hfill
    \begin{minipage}[t]{0.58\textwidth}
        \centering
        \includegraphics[width=\linewidth]{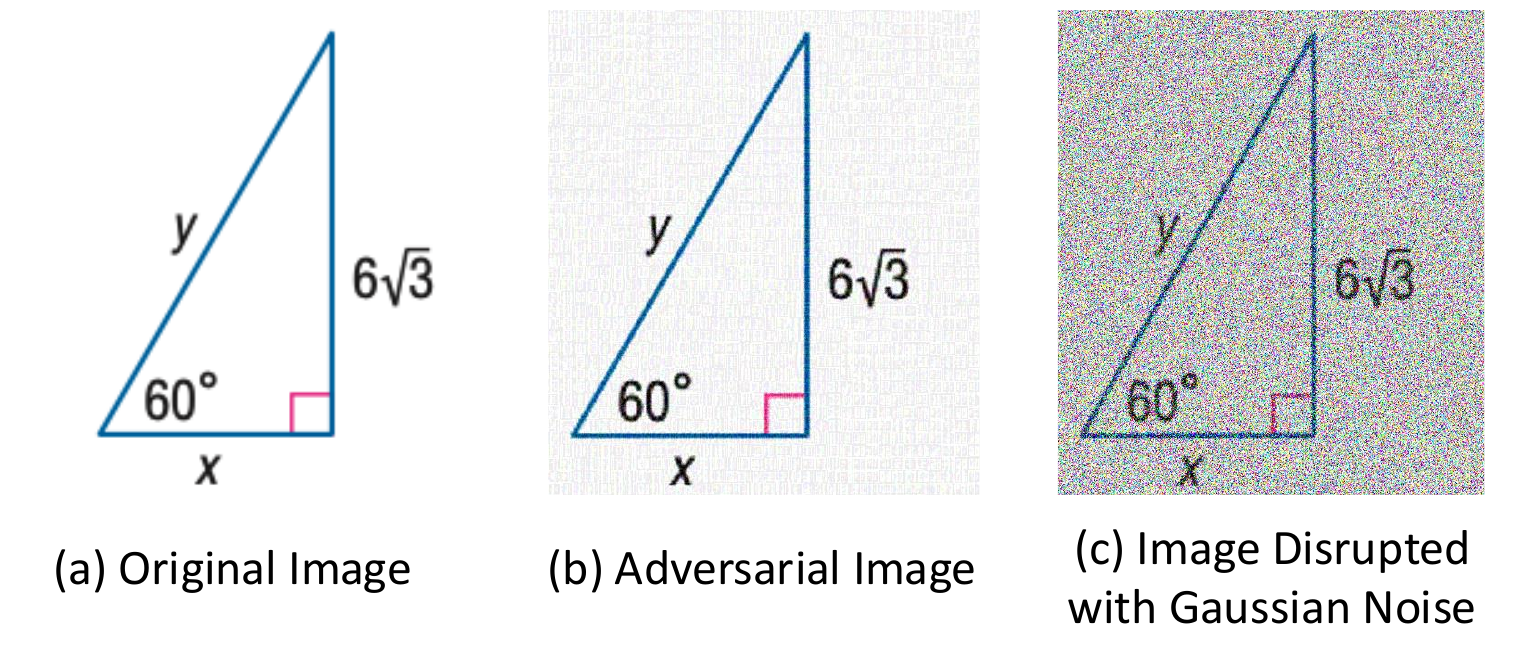}
	\caption{Visualization of an original image, its adversarial version (1 adversarial attack iteration with step size $\alpha=3/255$), and Gaussian noise (500-step) disrupted version. }
	\label{images}
    \end{minipage}
\end{figure}

\subsection{Ablation Studies}
We conduct ablation studies on Geometry3K with the group size $n=8$ and report in-domain accuracy. We run each experiment three times.

\noindent\textbf{Component analysis of SaEI.} Our SaEI consists of two main components, i.e., entropy-guided adversarial sampling
(EgAS), and token-selective entropy computation (TsEC). For the experiment with only EgAS, we use all tokens of each sampled response to compute an adversarial objective whose gradient is then back-propagated to the visual input. Sampling with adversarial samples enables the policy model to explore a larger response space. As shown in Tab.\ref{mainab}, after applying EgAS to vanilla GRPO, the performance is improved by 1.39$\%$ due to the improved policy exploration. Moreover, we use TsEC to discard the tokens with the lowest entropy to avoid distorting factual knowledge and those with the highest entropy, which already provide sufficient exploration. The TsEC results in a more substantial improvement over vanilla GRPO, with gains of up to 2.55$\%$.

\noindent\textbf{Comparison with random noise.}
We compare our SaEI with random noise, which is created by adding multiple-step Gaussian noise to the image, as suggested by NoiseRollout \cite{liu2025noisyrollout}.
As shown in Fig.\ref{random_noise}, 300-step Gaussian noise leads to the best result (53.41$\%$) among experiments with random noise. However, its result is 
1.61$\%$ lower than SaEI, which demonstrates the superiority of SaEI. We can also observe that too weak random noise, i.e., 200-step Gaussian noise can hardly improve the result of GRPO due to its inability to intervene in policy entropy. However, too strong random noise, i.e., 500-step Gaussian noise leads to an accuracy drop up to 2.75$\%$. The reason is that too strong random noise distorts the distribution of original images too much as shown in Fig.\ref{images} (c), thus impeding the learning from original images. In contrast, our SaEI can effectively intervene in policy entropy without substantially distorting the original image, as shown in Fig.\ref{images} (b). The inefficiency of random noise comes from lacking explicit relation with policy entropy. In contrast, our SaEI directly links policy entropy and the model input through adversarial attack.

\noindent\textbf{Ablation on maximum-entropy objectives.}
Our method is a sampling-stage approximation to maximum-entropy RL. Adding entropy bonus to advantage and entropy regularization to policy loss are two approaches to realize maximum-entropy RL by explicitly adding the entropy term in Eq.~\eqref{eq:maxent_objective}. However, explicitly adding the entropy term may bias or destabilize policy optimization. Entropy Adv.  \cite{cheng2026reasoning} adds entropy bonus to advantage. As shown in Tab.~\ref{tab:mm-eureka} and Tab.~\ref{tab:geo3k}, although it outperforms vanilla GRPO, it underperforms our method with a large gap. As shown in Tab.~\ref{tab:me}, we also add entropy regularization to policy loss with different weights $\alpha$.  A small $\alpha$ may provide insufficient entropy regularization, whereas a large $\alpha$ may bias policy optimization. Although $\alpha=1 \times 10^{-3}$ improves vanilla GRPO, it still lags behind our method.

\begin{table}[t]
    \centering
    \caption{Ablation on Maximum-entropy Objectives.}
    \label{tab:me}
    \resizebox{\linewidth}{!}{
    \begin{tabular}{lccccc}
    \toprule
    Method
    & Vanilla GRPO
    & $\alpha = 5 \times 10^{-4}$
    & $\alpha = 1 \times 10^{-3}$
    & $\alpha = 5 \times 10^{-3}$
    & Ours \\
    \midrule
    Accuracy (\%)
    & 52.47 $\pm$ 0.97
    & 52.41 $\pm$ 0.17
    & 53.24 $\pm$ 1.09
    & 49.23 $\pm$ 1.03
    & \textbf{55.02 $\pm$ 0.10} \\
    \bottomrule
    \end{tabular}
    }
  \end{table}

\paragraph{Ablation on entropy bounds in TsEC.}
  We ablate different entropy-bound strategies in TsEC, as shown in
  Table~\ref{tab:tsec_bound_ablation}. Without entropy bounds, the accuracy is
  $53.86$, indicating that using all tokens introduces noisy perturbation signals.
  Applying an upper bound to positive-advantage tokens improves the result to
  $54.85$, supporting our claim that high-entropy tokens in successful
  trajectories should be protected from further perturbation. The sign-aware
  strategy, which uses an upper bound for positive-advantage tokens and a lower
  bound for negative-advantage tokens, achieves the highest accuracy of $55.08$. The bilateral entropy bound obtains a
  competitive accuracy of $55.02$ with a negligible margin to $55.08$, so we adopt it as the default setting for its simplicity.

  \begin{table}[t]
    \centering
    \caption{Ablation study on different entropy-bound strategies in TsEC.}
    \label{tab:tsec_bound_ablation}
    \resizebox{\linewidth}{!}{
    \begin{tabular}{m{2.4cm}<{\centering} m{2.4cm}<{\centering} m{3.0cm}<{\centering} m{3.8cm}<{\centering}
  m{2.4cm}<{\centering}}
    \toprule
    Method
    & No bound
    & Upper for $A^+$
    & \shortstack[c]{Upper for $A^+$\\Lower for $A^-$}
    & Bilateral \\
    \midrule
    Accuracy (\%)
    & 53.86 $\pm$ 0.53
    & 54.85 $\pm$ 0.69
    & \textbf{55.08 $\pm$ 0.86}
    & 55.02 $\pm$ 0.10 \\
    \bottomrule
    \end{tabular}
    }
  \end{table}

\section{Conclusion}
We propose a sampling-stage approximation to entropy-regularized exploration by applying adversarial learning to entropy intervention in VLM reinforcement learning. We show that entropy-driven adversarial samples can effectively improve policy exploration in VLM reinforcement learning. To this end, we propose Selective-adversarial Entropy Intervention (SaEI) to achieve entropy intervention in RL sampling, instead of previous studies that simply control the update of specific tokens during policy optimization of RL. SaEI contains the entropy-guided adversarial sampling (EgAS) that enables the policy model to explore a larger answer space by formulating the policy entropy as an adversarial objective and attacking the visual input. Then, token-selective entropy computation (TsEC) is proposed to maximize the effectiveness of adversarial attack in EgAS via bilateral entropy bound. Extensive experiments show that our proposed EgAS can improve the generalization and robustness of current VLMs, leading to SOTA performance across multiple visual reasoning benchmarks.

\newpage
  {
    \small
    \bibliographystyle{plainnat}
    \bibliography{main}
  }

\newpage
\appendix

\section{Quantitative Connection Between $\epsilon$ and $\alpha$}
  \label{app:maxent_egas}

 This appendix provides the detailed derivations for the quantitative connection between the adversarial step
  size $\epsilon$ used in EgAS and the entropy regularization strength $\alpha$ in the maximum-entropy policy
  improvement objective. The connection is local and token-wise. It becomes exact when the adversarial entropy gradient acts as a logit-flattening direction.

  \subsection{Closed-Form Maximum-Entropy Policy}
  \label{sec:closed_form_maxent}

  Fix a token state $h_t=(I,Q,y_{<t})$, and let
  \begin{equation}
  p_0(a)
  =
  \pi_{\theta_{\mathrm{old}}}(a\mid h_t)
  =
  \mathrm{softmax}(z)(a)
  \label{eq:old_token_policy}
  \end{equation}
  be the old token policy with logits $z(a)$. Consider the local maximum-entropy improvement problem
  \begin{equation}
  p_\alpha^\star
  =
  \arg\max_{p\in\Delta}
  \left\{
  \mathbb{E}_{a\sim p}[r_t(a)]
  -
  \beta \mathrm{KL}(p\|p_0)
  +
  \alpha H(p)
  \right\},
  \label{eq:local_maxent_policy}
  \end{equation}
  where $\Delta$ is the probability simplex over the vocabulary, $r_t(a)$ is a token-level reward or action-value
  surrogate, and $\beta>0$ is the strength of the reference-policy constraint. The closed-form solution is
  \begin{equation}
  p_\alpha^\star(a)
  =
  \frac{
  p_0(a)^{\frac{\beta}{\beta+\alpha}}
  \exp\left(\frac{r_t(a)}{\beta+\alpha}\right)
  }{
  Z_\alpha
  },
  \qquad
  Z_\alpha
  =
  \sum_{a'}
  p_0(a')^{\frac{\beta}{\beta+\alpha}}
  \exp\left(\frac{r_t(a')}{\beta+\alpha}\right).
  \label{eq:closed_form_policy}
  \end{equation}
  Therefore, when the sampling-stage reward is absent or locally ignored, i.e., $r_t\equiv 0$,
  \begin{equation}
  p_\alpha^\star(a)
  =
  \mathrm{softmax}(\tau_\alpha z)(a),
  \qquad
  \tau_\alpha
  =
  \frac{\beta}{\beta+\alpha}.
  \label{eq:reward_free_flattening}
  \end{equation}
  Thus, in the reward-free sampling stage, maximum-entropy regularization is equivalent to flattening the old
  logits by the factor $\tau_\alpha=\beta/(\beta+\alpha)$. Larger $\alpha$ means smaller $\tau_\alpha$, hence a
  higher-entropy token distribution.

  \paragraph{Derivation of Eq.~\eqref{eq:closed_form_policy}.}
  The Lagrangian of Eq.~\eqref{eq:local_maxent_policy} is
  \begin{equation}
  \mathcal{L}(p,\eta)
  =
  \sum_a p(a)r_t(a)
  -
  \beta\sum_a p(a)\log\frac{p(a)}{p_0(a)}
  -
  \alpha\sum_a p(a)\log p(a)
  +
  \eta\left(\sum_a p(a)-1\right).
  \label{eq:local_maxent_lagrangian}
  \end{equation}
  Setting $\partial \mathcal{L}/\partial p(a)=0$ gives
  \begin{equation}
  r_t(a)
  +
  \beta\log p_0(a)
  -
  (\beta+\alpha)\log p(a)
  -
  (\beta+\alpha)
  +
  \eta
  =
  0.
  \label{eq:local_maxent_stationary}
  \end{equation}
  Solving for $p(a)$ and normalizing over actions yields Eq.~\eqref{eq:closed_form_policy}.

  \subsection{EgAS as a Local Exponentiated Policy Tilt}
  \label{sec:egas_policy_tilt}

  Let
  \begin{equation}
  F(I)
  =
  H_{S_\gamma}(\pi_{\theta_{\mathrm{old}}};(I,Q)),
  \qquad
  u
  =
  \mathrm{sign}(\nabla_I F(I)).
  \label{eq:egas_entropy_direction}
  \end{equation}
  For a small adversarial step $I_{\mathrm{adv}}=I+\epsilon u$, linearize the token logits:
  \begin{equation}
  z_\epsilon(a)
  =
  z(a)
  +
  \epsilon b(a)
  +
  O(\epsilon^2),
  \qquad
  b(a)
  =
  u^\top\nabla_I z(a).
  \label{eq:egas_logit_linearization}
  \end{equation}
  Ignoring $O(\epsilon^2)$ terms, the adversarial rollout policy is
  \begin{equation}
  p_\epsilon(a)
  =
  \mathrm{softmax}(z+\epsilon b)(a)
  =
  \frac{
  p_0(a)\exp(\epsilon b(a))
  }{
  \mathbb{E}_{a'\sim p_0}[\exp(\epsilon b(a'))]
  }.
  \label{eq:egas_exponentiated_tilt}
  \end{equation}
  Hence, EgAS is exactly an exponentiated policy tilt of the old policy under the first-order logit
  approximation.

  \begin{remark}
  Eq.~\eqref{eq:egas_exponentiated_tilt} is itself the closed-form solution of
  \begin{equation}
  \arg\max_{p\in\Delta}
  \left\{
  \epsilon\mathbb{E}_{a\sim p}[b(a)]
  -
  \mathrm{KL}(p\|p_0)
  \right\}.
  \label{eq:egas_kl_improvement}
  \end{equation}
  Thus, even without the logit-flattening condition below, EgAS can be interpreted as a KL-regularized soft
  policy improvement step with adversarial entropy-gradient pseudo-reward $b(a)$.
  \end{remark}

  \subsection{Exact Equivalence Under Logit Flattening}
  \label{sec:logit_flattening_equivalence}

  Comparing Eq.~\eqref{eq:reward_free_flattening} and Eq.~\eqref{eq:egas_exponentiated_tilt}, EgAS is exactly
  equivalent to the reward-free closed-form maximum-entropy solution if and only if, up to an action-independent
  constant,
  \begin{equation}
  z(a)
  +
  \epsilon b(a)
  =
  \frac{\beta}{\beta+\alpha}z(a)
  +
  c.
  \label{eq:equivalence_logit_condition}
  \end{equation}
  Equivalently,
  \begin{equation}
  \epsilon b(a)
  =
  -
  \frac{\alpha}{\beta+\alpha}z(a)
  +
  c.
  \label{eq:equivalence_logit_condition_rearranged}
  \end{equation}
  Since action-independent constants do not change a softmax distribution, this condition says that the
  adversarial entropy gradient must flatten the centered logits. Suppose, for the selected token state, the
  adversarial direction satisfies
  \begin{equation}
  b(a)
  =
  -\kappa
  \left(
  z(a)
  -
  \mathbb{E}_{a'\sim p_0}[z(a')]
  \right),
  \qquad
  \kappa>0.
  \label{eq:logit_flattening_condition}
  \end{equation}
  Then EgAS and the closed-form reward-free maximum-entropy policy are identical when
  \begin{equation}
  \kappa\epsilon
  =
  \frac{\alpha}{\beta+\alpha}.
  \label{eq:kappa_epsilon_alpha_relation}
  \end{equation}
  Solving for $\alpha$ yields the quantitative connection
  \begin{equation}
  \alpha_{\mathrm{eff}}
  =
  \frac{\beta\kappa\epsilon}{1-\kappa\epsilon},
  \qquad
  0\le \kappa\epsilon < 1.
  \label{eq:alpha_eff_exact}
  \end{equation}
  For small adversarial steps, this becomes the first-order relation
  \begin{equation}
  \alpha_{\mathrm{eff}}
  =
  \beta\kappa\epsilon
  +
  O(\epsilon^2).
  \label{eq:alpha_eff_first_order}
  \end{equation}

  \paragraph{Estimating $\kappa$ from model quantities.}
  The exact flattening condition in Eq.~\eqref{eq:logit_flattening_condition} need not hold perfectly. A data-
  dependent effective flattening coefficient can be obtained by projecting the adversarial logit direction $b$
  onto the centered negative logit direction. Define
  \begin{equation}
  \widetilde{z}(a)
  =
  z(a)
  -
  \mathbb{E}_{a\sim p_0}[z(a)],
  \qquad
  \widetilde{b}(a)
  =
  b(a)
  -
  \mathbb{E}_{a\sim p_0}[b(a)].
  \label{eq:centered_logit_direction}
  \end{equation}
  Then
  \begin{equation}
  \kappa_{\mathrm{proj}}
  =
  -
  \frac{
  \mathbb{E}_{a\sim p_0}[\widetilde{b}(a)\widetilde{z}(a)]
  }{
  \mathbb{E}_{a\sim p_0}[\widetilde{z}(a)^2]
  }.
  \label{eq:kappa_projection}
  \end{equation}
  Substituting $\kappa_{\mathrm{proj}}$ into Eq.~\eqref{eq:alpha_eff_exact} gives a measurable token-wise
  effective entropy coefficient:
  \begin{equation}
  \alpha_{\mathrm{eff},t}(\epsilon)
  =
  \frac{
  \beta\kappa_{\mathrm{proj},t}\epsilon
  }{
  1-\kappa_{\mathrm{proj},t}\epsilon
  }.
  \label{eq:tokenwise_alpha_eff}
  \end{equation}
  For a selected token set $S_\gamma$, one may report
  \begin{equation}
  \overline{\alpha}_{\mathrm{eff}}
  =
  \frac{1}{|S_\gamma|}
  \sum_{t\in S_\gamma}
  \alpha_{\mathrm{eff},t}.
  \label{eq:average_alpha_eff}
  \end{equation}

  \subsection{Entropy-Gain Calibration}
  \label{sec:entropy_gain_calibration}

  The equivalence above matches the full token distribution. A weaker but often more robust calibration matches
  the first-order entropy gain. Under Eq.~\eqref{eq:reward_free_flattening}, the derivative of token entropy at $
  \alpha=0$ is
  \begin{equation}
  \left.
  \frac{\partial}{\partial\alpha}
  H\left(
  \mathrm{softmax}\left(
  \frac{\beta}{\beta+\alpha}z
  \right)
  \right)
  \right|_{\alpha=0}
  =
  \frac{1}{\beta}
  \mathrm{Var}_{a\sim p_0}[z(a)].
  \label{eq:entropy_derivative_alpha}
  \end{equation}
  For the selected average entropy $F(I)$, EgAS gives
  \begin{equation}
  F(I+\epsilon u)-F(I)
  =
  \epsilon\|\nabla_I F(I)\|_1
  +
  O(\epsilon^2).
  \label{eq:egas_entropy_gain}
  \end{equation}
  Let
  \begin{equation}
  V_S
  =
  \frac{1}{|S_\gamma|}
  \sum_{t\in S_\gamma}
  \mathrm{Var}_{a\sim p_{0,t}}[z_t(a)].
  \label{eq:selected_variance}
  \end{equation}
  Equating the two first-order entropy gains gives
  \begin{equation}
  \alpha_{\mathrm{eff}}
  \approx
  \beta\epsilon
  \frac{
  \|\nabla_I F(I)\|_1
  }{
  V_S
  },
  \qquad
  V_S>0.
  \label{eq:entropy_gain_alpha_eff}
  \end{equation}
  This formula is useful empirically because all quantities on the right-hand side can be computed during the
  EgAS attack.

  \begin{remark}
  Eq.~\eqref{eq:alpha_eff_exact} is a distribution-level equivalence under a logit-flattening condition.
  Eq.~\eqref{eq:entropy_gain_alpha_eff} is a weaker entropy-gain matching rule that remains meaningful even when
  the adversarial direction is not perfectly aligned with the negative centered logits.
  \end{remark}

\section{Training Dynamics of SaEI and GRPO}
\begin{figure}[t]
	\centering
	\includegraphics[width=0.7\textwidth]{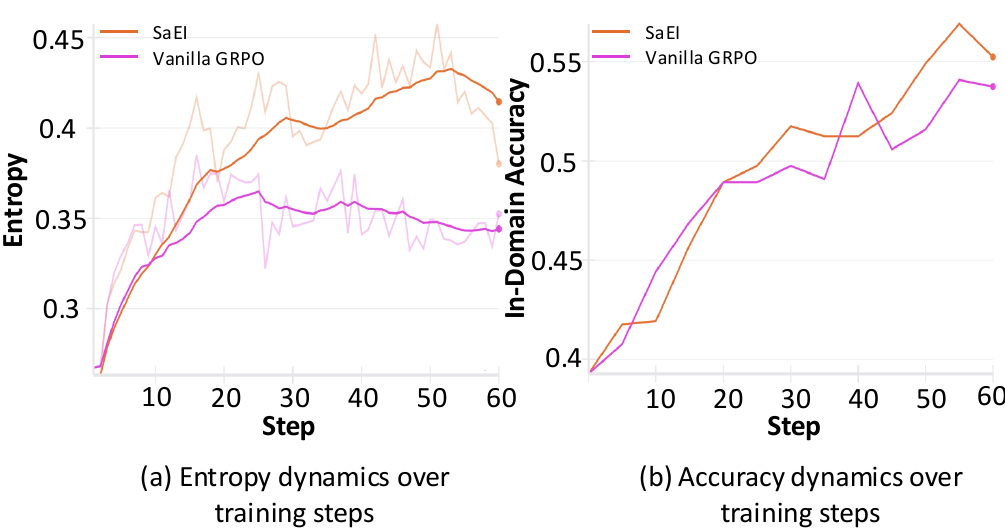}
	\caption{Comparison of our SaEI and vanilla GRPO in terms of entropy dynamics and accuracy dynamics over training steps. The dark curves in (a) show data processed with an exponential moving average (EMA), while the light lines show the original data. }
	\label{moti1}
\end{figure}

We visualize the training dynamics of our SaEI and vanilla GRPO in Fig.~\ref{moti1}. Our SaEI can improve policy exploration by adequately increasing policy entropy during training. With better policy exploration, SaEI substantially outperforms vanilla GRPO.

\section{Additional Ablation Studies}

\begin{figure}[t]
	\centering
	\includegraphics[width=0.7\textwidth]{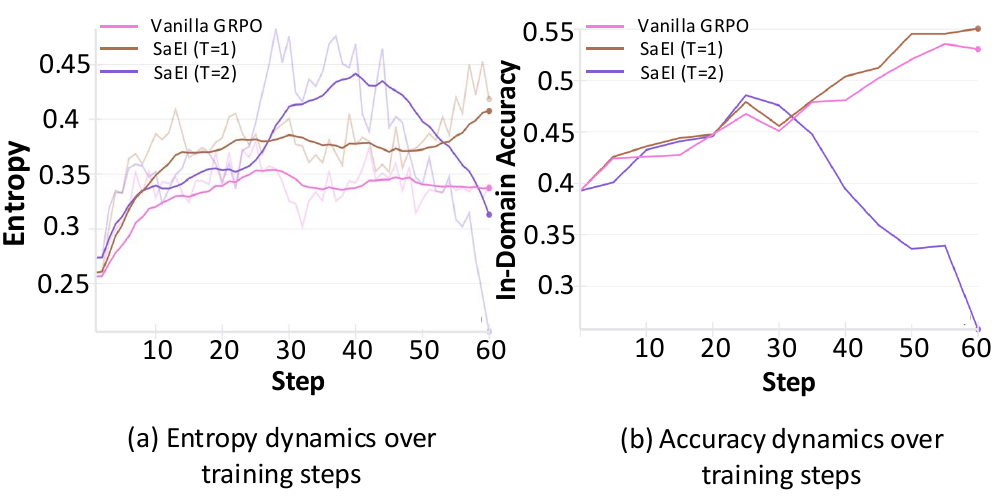}
	\caption{Ablation on iteration number of adversarial attack.}

	\label{2t}
\end{figure}

\noindent\textbf{Ablation on the number of adversarial attack iterations.} Our SaEI takes $T=1$ adversarial attack iteration, and we compare it with $T=2$. As shown in Fig.~\ref{2t}, $T=2$ causes a rapid entropy increase and a rapid accuracy decline after 30 steps. Since a larger $T$ leads to training instability, we set $T$ to 1, which is also more computationally efficient.

\noindent\textbf{Ablation on $\gamma$ of bilateral bound.}
We ablate the entropy-selection ratio $\gamma$ in TsEC. The ``No bound'' variant applies the entropy objective
  without filtering tokens by entropy, while $\gamma$ controls the bilateral bound for selecting moderate-entropy tokens. As shown in Tab.~\ref{tab:gamma_ablation}, using the entropy bound improves over the no-bound baseline
  across most settings, with $\gamma=0.33$ achieving the best accuracy of $55.02\pm0.10$. A smaller bound ratio
  still provides consistent gains, whereas $\gamma=0.40$ leads to degraded performance. We therefore set $\gamma=0.33$
  by default.
\begin{table}[t]
      \centering
      \caption{Ablation on different values of $\gamma$.}
      \label{tab:gamma_ablation}
      \resizebox{\linewidth}{!}{
      \begin{tabular}{lccccc}
      \toprule
      Method
      & No bound
      & $\gamma = 0.10$
      & $\gamma = 0.20$
      & $\gamma = 0.33$
      & $\gamma = 0.40$ \\
      \midrule
      Accuracy (\%)
      & 53.86 $\pm$ 0.53
      & 54.47 $\pm$ 1.62
      & 54.63 $\pm$ 0.51
      & \textbf{55.02 $\pm$ 0.10}
      & 53.63 $\pm$ 1.25 \\
      \bottomrule
      \end{tabular}
      }
  \end{table}

\section{Pseudocode of Our Method}
\label{sec:intro}
Algorithm \ref{alg:train} presents a simplified overview to better illustrate the workflow of our proposed method.

\begin{algorithm}[htb]
   \caption{Selective-adversarial Entropy Intervention}
   \label{alg:train}
    \begin{algorithmic}[]
   \STATE {\bfseries Input:} Policy model $\pi_\theta$, old model $\pi_{\theta_{old}}$, reference model $\pi_{\theta_{ref}}$, clean responses number $n_1$, adversarial responses number $n_2$, clip parameter $\epsilon$, adversarial attack iteration number $S$, adversarial attack step size $\eta_{\mathrm{adv}}$, training steps $T_{training}$
   
   \FOR{$t=1,\cdots, T_{traing}$}
        \STATE \textbf{Sample} batch $(I, Q) \sim P_D$
        \STATE \textbf{Sample} responses $\{y^i\}_{i=1}^{n_1}$ from $\pi_{\theta_{old}}(y | I, Q)$
        \STATE \textbf{Compute} entropy $\mathcal{H}(\pi_{\theta_{old}}, \{y^i\}_{i=1}^n )
  = - \frac{1}{n_1} \sum_{i=1}^{n_1} \frac{1}{|y^i|} \sum_{t=1}^{|y^i|} \mathbb{E}_{y_t^i \sim \pi_{\theta_{old}}} [\log \pi_{\theta_{\text{old}}}(y_t^i|q, y_{<t}^i) ]$
        \STATE \textbf{Rank} and \textbf{select} $\mathcal{H}(\pi_{\theta_{old}}, \{y^i\}_{i=1}^{n_1} ) = - \frac{1}{n_1} \sum_{i=1}^{n_1} \frac{1}{|\hat{y}^i|} \sum_{t=1}^{|\hat{y}^i|} \mathbb{E}_{\hat{y}_t^i \sim \pi_{\theta_{old}}} [\log \pi_{\theta_{\text{old}}}(\hat{y}_t^i|q, y_{<t}^i) ],$  
        \STATE $\text{s.t.} \quad \frac{1}{3} * |y^i|<rank(\hat{y}_t^i) <\frac{2}{3} *  |y^i|$
        \STATE \textbf{Initialize} $I^{1}_{adv}=I$
        \FOR{$s=1,\cdots, S$}
            \STATE \textbf{PGD Attack} $I^{s+1}_{adv}
 = I^{s}_{adv} + \alpha \cdot sign(\nabla_{I^{s}_{adv}}-\mathcal{H}(\pi_{\theta_{old}}, \{y^i\}_{i=1}^{n_1})),$
            \ENDFOR
        \STATE \textbf{Sample} responses $\{y^i\}_{i=1}^{n_2}$ from $\pi_{\theta_{old}}(y | I_{adv}, Q)$
        \STATE \textbf{Compute} rewards $\{R^i\}_{i=1}^{n_1+n_2} = \{ \{R^i\}_{i=1}^{n_1} \cup \{R^i\}_{i=1}^{n_2}\}$
        \STATE \textbf{Compute} advantages $\tilde{A}^i_t
 = \frac{R^i-mean(\{R^i\}_{i=1}^{n_1+n_2})}{std(\{R^i\}_{i=1}^{n_1+n_2})}$
        \STATE \textbf{Compute} GRPO objective $\mathcal{J}(\theta) = \mathbb{E} \frac{1}{{n_1+n_2}} \sum_{i=1}^{n_1+n_2} \frac{1}{|y^i|} \sum_{t=1}^{|y^i|} \Big \{ \min \Big[ \frac{\pi_{\theta}(y_t^i|q, y_{<t}^i)}{\pi_{\theta_{\text{old}}}(y_t^t|q , y_{<t}^i)} \tilde{A}^i_t,$  
        \STATE $\text{clip}\Big(\frac{\pi_{\theta}(y_t^i|q, y_{<t}^i)}{\pi_{\theta_{\text{old}}}(y_t^i|q, y_{<t}^i)}, 1-\epsilon, 1+\epsilon \Big) \tilde{A}^i_t  \Big] - \beta D_{kl}  \Big \}$
        \STATE \textbf{Update} policy model $\theta$ by descending $-\nabla_{\theta}J(\theta)$.
        \STATE \textbf{Update} old model $\theta_{old} = \theta$
    \ENDFOR
    \end{algorithmic}
\end{algorithm}

\section{Reasoning Prompt}
We use the following reasoning prompt from EasyR1 to stimulate the reasoning process of VLMs.

\begin{figure}[htb]
	\centering
	\includegraphics[width=0.47\textwidth]{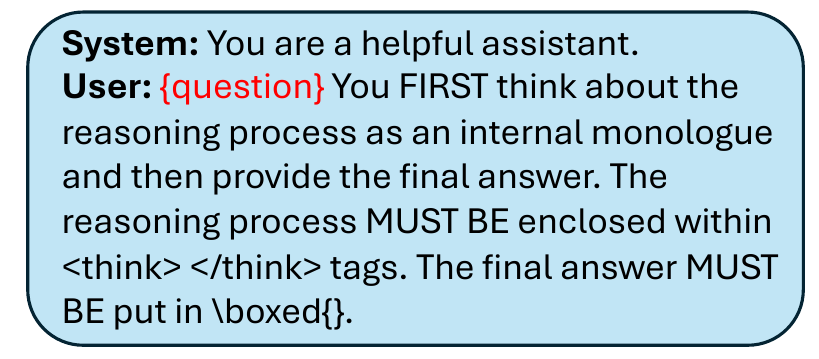}
	\caption{Reasoning prompt used to stimulate the reasoning process of VLMs.}

	\label{tsec}
\end{figure}

\section{Key Implementation Details}
In this section, we summarize key implementation details of our method.

\begin{table}[htb]
\center
\caption{Key Implementation Details.}
\begin{tabular}{cc}
\hline
\multicolumn{1}{l}{\textbf{Parameter}} & \textbf{Value}         \\ \hline
\multicolumn{2}{c}{\textbf{All experiments}}                    \\ \hline

Global Batch Size                      & 128                    \\
Rollout Batch Size                     & 512                    \\
Optimizer                              & AdamW                  \\
Learning Rate                          & 1e-6                   \\
Policy Loss Aggregation                & $token-mean$           \\
Adversarial attack iteration number    & 1                      \\ 
Sampled response number                & $n_1=n_2=6$            \\ \hline
\multicolumn{2}{c}{\textbf{Experiments on Geometry3K}}         \\ \hline
Total Optimization Steps               & 60                     \\
Adversarial attack step size $\eta_{\mathrm{adv}}$  & 3/255                 \\ \hline
\multicolumn{2}{c}{\textbf{Experiments on MM-Eureka}}           \\ \hline
Total Optimization Steps               & 90                     \\
Adversarial attack step size $\eta_{\mathrm{adv}}$  & 2/255                 \\ \hline
\end{tabular}
\label{geosk_n8}
\end{table}

\section{Case Studies}
\begin{figure*}[h]
	\centering
	\includegraphics[width=0.95\textwidth]{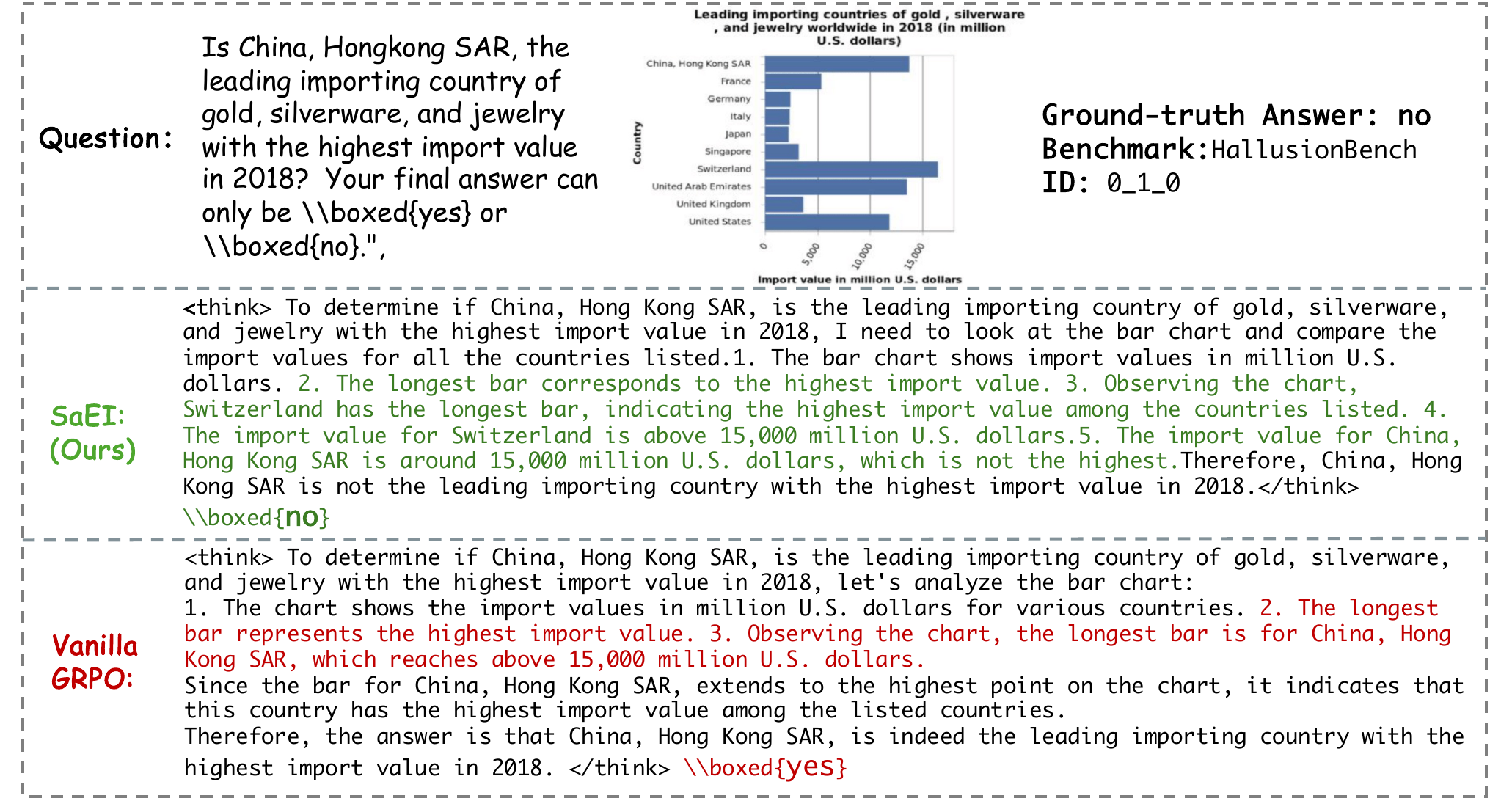}
	\caption{An example of a visual question-answer pair. The sample comes from HallusionBench, an OOD benchmark.}
	\label{case}
\end{figure*}

In Fig.\ref{case}, we present an example of visual question-answer pair. SaEI and vanilla GRPO are both trained on MM-Eureka dataset. Fig.\ref{case} shows that even trained with a mathematics reasoning dataset, our SaEI model can provide correct reasoning processes and answers to general visual questions, even in cases where vanilla GRPO fails.

\begin{figure*}[t]
	\centering
	\includegraphics[width=0.97\textwidth]{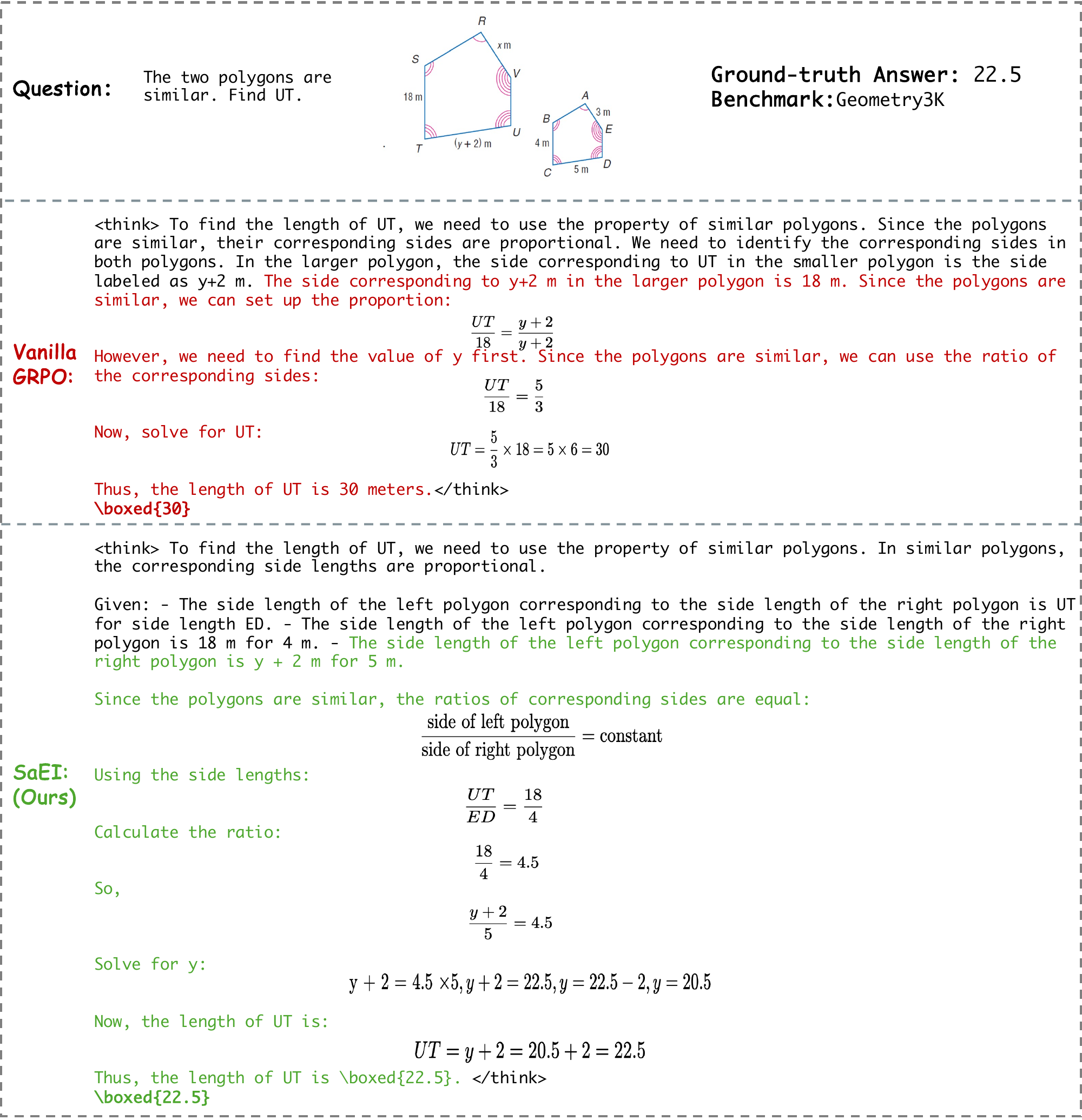}
	\caption{Vanilla GRPO gives a wrong response while our SaEI gives a correct response.}
	\label{case1}
\end{figure*}

Fig.\ref{case1} shows another case of geometry problems. SaEI and vanilla GRPO are trained on the Geometry3K dataset. As shown in Fig.\ref{case1}, vanilla GRPO gives a wrong response while our SaEI gives a correct response. Although vanilla GRPO realizes the property of similar polygons, it fails to find the correct side corresponding to ST and mistakes AE for BC in the smaller polygon. Although SaEI gives a correct response, it mistakenly writes ED for CD when writing the expression. Nonetheless, SaEI uses the value of CD, leading to a correct response. Due to the fact that we adopt an outcome-based reward function by following GRPO, which applies no constraint to the reasoning process, it is possible to obtain correct responses even with an imperfect reasoning process.

\section{Limitations}
  \label{sec:limitations}

  The main limitation of SaEI is its additional computational cost during RL training. Unlike vanilla GRPO, SaEI
  performs entropy-guided adversarial sampling by first generating clean responses, computing the selective token
  entropy, back-propagating the entropy objective to the visual input, and then sampling additional responses
  from the adversarial visual context. Therefore, SaEI introduces both an extra backward pass for constructing
  the adversarial image and extra forward passes for adversarial rollouts. To mitigate this cost, we use a single adversarial attack iteration in all experiments. Nevertheless, SaEI
  remains more expensive than standard GRPO and entropy intervention methods that operate only during policy
  optimization. This overhead may become more pronounced for larger VLMs, longer responses, or larger rollout
  group sizes. Future work may improve efficiency by reducing the number of attacked samples, reusing
  activations, or approximating the entropy-gradient direction with cheaper surrogate objectives.

\newpage
\section*{NeurIPS Paper Checklist}

\begin{enumerate}

\item {\bf Claims}
    \item[] Question: Do the main claims made in the abstract and introduction accurately reflect the paper's contributions and scope?
    \item[] Answer: \answerYes{}.
    \item[] Justification: Please refer to our abstract and introduction.

\item {\bf Limitations}
    \item[] Question: Does the paper discuss the limitations of the work performed by the authors?
    \item[] Answer:  \answerYes{}.
    \item[] Justification: We discuss our limitations in our supplementary materials. Our limitations are mainly about increased training time incurred by our method.

\item {\bf Theory assumptions and proofs}
    \item[] Question: For each theoretical result, does the paper provide the full set of assumptions and a complete (and correct) proof?
    \item[] Answer: \answerYes{}.
    \item[] Justification: Please refer to our appendix A.

    \item {\bf Experimental result reproducibility}
    \item[] Question: Does the paper fully disclose all the information needed to reproduce the main experimental results of the paper to the extent that it affects the main claims and/or conclusions of the paper (regardless of whether the code and data are provided or not)?
    \item[] Answer:  \answerYes{}.
    \item[] Justification: Please refer to our Implementation Details section. Our code will be released after the acceptance of this paper.

\item {\bf Open access to data and code}
    \item[] Question: Does the paper provide open access to the data and code, with sufficient instructions to faithfully reproduce the main experimental results, as described in supplemental material?
    \item[] Answer: \answerNo{}.
    \item[] Justification: We plan to release our code after the acceptance of this paper.

\item {\bf Experimental setting/details}
    \item[] Question: Does the paper specify all the training and test details (e.g., data splits, hyperparameters, how they were chosen, type of optimizer) necessary to understand the results?
    \item[] Answer: \answerYes{}
    \item[] Justification: Please refer to our Implementation Details section.

\item {\bf Experiment statistical significance}
    \item[] Question: Does the paper report error bars suitably and correctly defined or other appropriate information about the statistical significance of the experiments?
    \item[] Answer: \answerYes{}.
    \item[] Justification: The paper reports mean and standard deviation over multiple runs for the main experimental results.

\item {\bf Experiments compute resources}
    \item[] Question: For each experiment, does the paper provide sufficient information on the computer resources (type of compute workers, memory, time of execution) needed to reproduce the experiments?
    \item[] Answer:  \answerYes{}.
    \item[] Justification: All our experiments are conducted with 8 H100 GPUs. And training steps are presented in Implementation Details.
    
\item {\bf Code of ethics}
    \item[] Question: Does the research conducted in the paper conform, in every respect, with the NeurIPS Code of Ethics \url{https://neurips.cc/public/EthicsGuidelines}?
    \item[] Answer: \answerYes{}.
    \item[] Justification: The research conforms to the NeurIPS Code of Ethics.

\item {\bf Broader impacts}
    \item[] Question: Does the paper discuss both potential positive societal impacts and negative societal impacts of the work performed?
    \item[] Answer: \answerYes{}
    \item[] Justification: The paper discusses the broader implications of improving VLM training methods, including potential benefits for model capability.
    
\item {\bf Safeguards}
    \item[] Question: Does the paper describe safeguards that have been put in place for responsible release of data or models that have a high risk for misuse (e.g., pre-trained language models, image generators, or scraped datasets)?
    \item[] Answer: \answerNA{}.
    \item[] Justification: The paper does not release data or models that pose a high risk for misuse.

\item {\bf Licenses for existing assets}
    \item[] Question: Are the creators or original owners of assets (e.g., code, data, models), used in the paper, properly credited and are the license and terms of use explicitly mentioned and properly respected?
    \item[] Answer: \answerYes{}.
    \item[] Justification: We cite the original sources of all datasets used in this work and describe their
  licenses and terms of use where available. All assets are used in accordance with their respective
  licenses.

\item {\bf New assets}
    \item[] Question: Are new assets introduced in the paper well documented and is the documentation provided alongside the assets?
    \item[] Answer: \answerNA{}.
    \item[] Justification: The paper does not release new assets.

\item {\bf Crowdsourcing and research with human subjects}
    \item[] Question: For crowdsourcing experiments and research with human subjects, does the paper include the full text of instructions given to participants and screenshots, if applicable, as well as details about compensation (if any)? 
    \item[] Answer: \answerNA{}.
    \item[] Justification: The paper does not involve crowdsourcing or research with human subjects; all
  evaluations are conducted automatically.

\item {\bf Institutional review board (IRB) approvals or equivalent for research with human subjects}
    \item[] Question: Does the paper describe potential risks incurred by study participants, whether such risks were disclosed to the subjects, and whether Institutional Review Board (IRB) approvals (or an equivalent approval/review based on the requirements of your country or institution) were obtained?
    \item[] Answer: \answerNA{}.
    \item[] Justification: The paper does not involve crowdsourcing or research with human subjects.

\item {\bf Declaration of LLM usage}
    \item[] Question: Does the paper describe the usage of LLMs if it is an important, original, or non-standard component of the core methods in this research? Note that if the LLM is used only for writing, editing, or formatting purposes and does \emph{not} impact the core methodology, scientific rigor, or originality of the research, declaration is not required.
    \item[] Answer: \answerYes{}.
    \item[] Justification: Please refer to our Implementation Details section.

\end{enumerate}

\end{document}